# Commentary on Kosmyna, N., Hauptmann, E., Yuan, Y. T., Situ, J., Liao, X. H., Beresnitzky, A. V., ... & Maes, P. (2025). Your Brain on ChatGPT: Accumulation of Cognitive Debt when Using an AI Assistant for Essay Writing Task. arXiv preprint arXiv:2506.08872

Miloš Stanković[1,2,*], Ella Hirche[2,+], Sarah Kollatzsch[2,+], Julia Nadine Doetsch[2]

[1]Department of Clinical and Health Psychology, Faculty of Psychology, University of Vienna, Vienna, Austria

[2]Behavioral Epidemiology, Institute of Clinical Psychology and Psychotherapy, Technische Universität Dresden, Germany

*Correspondence concerning this article should be addressed to Miloš Stanković, Department of Clinical and Health Psychology, Faculty of Psychology, University of Vienna, Vienna, Austria. Email: milos.stankovic@univie.ac.at

+equal contributors

Miloš Stanković: https://orcid.org/0000-0001-9767-5634;

Julia Nadine Doetsch: https://orcid.org/0000-0003-1388-9542

## Abstract

Recently published work titled *Your Brain on ChatGPT: Accumulation of Cognitive Debt When Using an AI Assistant for Essay Writing Task* by Kosmyna et al. (2025) has sparked a vivid debate on the topic of artificial intelligence (AI) and human performance. We sincerely congratulate Kosmyna et al. for initiating such important research, collecting a valuable dataset, and establishing highly automated pipelines for Natural Language Processing (NLP) analyses and scoring. **We aim to provide constructive comments** that may improve the manuscript's readiness for peer-reviewed publication, as some results by Kosmyna et al. (2025) could be interpreted more conservatively. Our primary concerns focus on: (i) study design considerations, including the limited sample size; (ii) the reproducibility of the analyses; (iii) methodological issues related to the EEG analysis; (iv) inconsistencies in the reporting of results; and (v) limited transparency in several aspects of the study's procedures and findings.

## Introduction

The preprint presents an extensive and ambitious multimethod investigation of LLM assistance in essay writing, examining neural activity, behavioral outcomes, and the linguistic properties of the resulting essays. The study included four sessions conducted over four months. Sessions 1–3 employed a between-subjects design. Participants were assigned to one of three fixed groups (LLM, Search Engine, or Brain-only) and used the same tool across all sessions. In Session 4, a within-participants design was introduced: participants from the LLM wrote without assistance (LLM-to-Brain), while Brain-only participants switched to LLM assistance (Brain-to-LLM). Fifty-four participants took part in Sessions 1–3 (18 per group), and 18 completed Session 4. Overall, the authors aimed to examine the cognitive cost of LLM use in an educational writing context, including neurophysiological and behavioral differences across groups and the effects of LLM use on memory and perceived ownership of the written essay.

**Kosmyna et al. (2025) reported significant group differences**, with linguistic patterns remaining stable within groups and EEG measures generally indicating reduced neural engagement as external support increased. However, closer examination of the data suggests greater neural engagement in the beta frequency band for the LLM condition compared to the brain-only condition (when counting significant connections, as performed in other experimental conditions), reducing the consistency of this finding. The authors further reported that LLM users exhibited the weakest functional connectivity, the lowest sense of ownership, and the poorest recall performance. These effects did not fully normalize following the tool switch in Session 4 for the LLM-to-Brain group. Based on converging neural, linguistic, and behavioral evidence, the authors concluded that prolonged LLM use may impair cognitive engagement and learning-related processes.

**Nevertheless, in our view, several methodological limitations** may have contributed to the reported (over)interpretations and therefore require further analysis.

**The key limitation is the relatively small sample size relative to the number of comparisons performed.** A basic a priori power analysis using G*Power (Faul et al., 2007) for a repeated-measures ANOVA ($f$ = 0.25, α = .05, power = .95, three groups, four measurements, $r$ = .50) indicates that approximately N = 159 participants would be required for adequate power. Thus, the power of the reported study is likely lower, especially for analyses involving Session 4 or analyses on essay topics. For example, several topic-specific figures were interpreted based on extremely small subsamples (sometimes only 2–4 essays per group), which is insufficient for identifying stable patterns—particularly in linguistic and NLP-based analyses characterized by high inter-text variability. Consequently, the study raises substantial concerns about statistical reliability, inflated effect estimates, and the interpretability of neural and behavioral differences. Low statistical power, whether due to a small sample size, small effects, or both, reduces the likelihood that a statistically significant result truly reflects a meaningful effect (Button et al., 2013).

**Moreover, beyond issues of power, several inconsistencies in reporting may further complicate the interpretation of the results**:

A critical inconsistency appears on p. 42 (Figure 12), where a metric is reported for groups that do not seem to align with the rest of the manuscript (e.g., "Brain-only-to-Search-Engine" group). Such discrepancies may raise concerns regarding selective reporting and limit the transparency of the analytic procedures.

Another inconsistency concerns the reporting of the Session 4 interview results. On page 37, the authors report higher quoting correctness for the LLM-to-Brain-only group compared to the Brain-only-to-LLM group, whereas in the discussion, the results are described in the opposite direction. This divergence in reporting can lead to significantly different interpretations of the data.

**The reproducibility of the analyses is also difficult to evaluate due to the limited methodological details provided in the preprint.** For example, the authors state that 55 participants completed all experiments, yet only 54 participants are reported in the analysis. The criteria for excluding the participant who completed all experimental sessions from the analysis remain unclear. Additionally, although numerous repeated-measures ANOVAs are reported, it is not specified whether each comparison was analyzed separately or if corrections for multiple comparisons were applied. F-statistics are often omitted, and p-values are presented without sufficient clarification regarding their derivation. The criteria for defining "significant connections" in the EEG analyses are also ambiguous, with no clear explanation of whether these connections reflect temporal changes, group differences, or a combination of both. These gaps in reporting hinder independent assessment of the analytic validity of the neural and behavioral findings.

**There are multiple methodological issues regarding the EEG analysis, which constitutes a central component of the manuscript.** Although the authors report conducting statistical analyses, the procedures are insufficiently described, making it difficult to assess the validity of the reported findings. Based on the available information, it appears that a repeated-measures ANOVA (rmANOVA) (2/3 groups x 4 sessions) was performed per electrode pair. However, the authors limited the number of ANOVAs to 1000, without clearly

specifying the criteria used to select these analyses. Additionally, the results of these ANOVAs are not reported, and it remains unclear whether the authors focused on initially significant connections in the ANOVA or considered other connections.

Instead, the authors reported results from comparisons between single groups or sessions, suggesting the use of post-hoc tests. False Discovery Rate (FDR) correction was applied "when multiple comparisons were involved" (p. 78). However, this phrasing raises **concerns regarding the adequacy of the statistical corrections**, particularly given that up to 1000 ANOVAs were conducted in addition to the post-hoc tests. The authors neither stated the FDR level used nor reported whether p-values of single connections were compared to a predetermined significance threshold. Instead, p-values were only compared to the alpha level of 0.05, which undermines confidence in the consistent use of FDR correction. Thus, the statistical validity of the reported results remains uncertain.

Further, in the context of the study, a "significant connection" was defined as a directed statistical coupling between two electrodes that was significantly stronger in one group than in another. Therefore, **no direct conclusions could be drawn regarding the absolute strength of the connections.** Thus, a group with fewer significant connections should not necessarily be interpreted as exhibiting weaker overall neural activity. This important distinction could have been more clearly emphasized to reduce the risk of overinterpretation. Moreover, instead of categorizing connections as "highly significant" versus "only significant", the use of effect sizes would have provided a more valid interpretation, increasing the overall strength.

**Furthermore, some findings may be overinterpreted.** For example, certain between-group comparisons are discussed without statistical testing or confidence intervals. Conceptual claims—such as a "diminished sense of cognitive agency" or reduced "value" of LLM-assisted essays—are not adequately supported by the study's operationalizations. Specifically, the interview questions assessed perceived authorship rather than cognitive agency or the intrinsic value of the work. Consequently, the observed differences between participants may also reflect actual differences in what they objectively created themselves as opposed to a subjective feeling of agency. Moreover, **the interview data were reported only for a subset of participants** (p. 32-33), without justification for this selective reporting. Additionally, the claim that the Brain-only group exhibits the strongest neural connectivity oversimplifies the pattern of the results. In fact, the LLM group demonstrated stronger connectivity than the Web-Engine group in certain frequency bands, indicating a more nuanced neural engagement. These examples highlight a **tendency to selectively emphasize findings that align with the authors' hypotheses.** This is also apparent in the limited discussion of the Brain-Only group's underperformance in certain metrics, such as the satisfaction metric. Notably, in Session 1, satisfaction ratings in the Brain-only group were lower than in other groups, and although satisfaction levels increased over time, they never reached the levels observed in the Search Engine group. However, the discussion states that "participants who were in the Brain-only group reported higher satisfaction" (p. 143), which does not fully align with the reported data.

One of the key behavioural markers the authors used in their argumentation was "correct quoting". To enhance the reproducibility of the study, **the authors should provide a more detailed explanation of how participant citations were matched to essay quotations.** Furthermore, confounding variables such as sentence complexity, which could be higher in LLM-generated content, should be considered. Finally, it is questionable whether reciting a single statement from the essays shows learning and memory encoding processes of interest, as semantic understanding seems more relevant. Although the authors indicated that participants were asked about their argumentation, which could have provided a more robust operationalization of cognitive engagement, no results of this inquiry were reported. Consequently, the "correct quoting" metric, as presented by the authors, does not strongly support the claim of decreased cognitive engagement to the extent implied.

**Future research could benefit from addressing three additional design considerations,** alongside those  proposed by the authors**:**

**First,** Session 4 effects should be interpreted cautiously, as repeated exposure to the same prompts may introduce familiarity or practice effects, potentially confounding neural or behavioral outcomes. Controlling for groups using the same tool on a familiar prompt could help eliminate these practice effects. Additional control variables, such as specific competencies, should also be considered due to their potential effects on cognitive engagement. The authors' proposal to adopt a longitudinal design is a promising direction. Allowing participants to use AI tools in a more naturalistic manner, particularly by providing sufficient time for essay writing, may yield more ecologically valid results. Involving control groups in the crossover phase, alongside relevant control variables, would increase confidence in the reported effects. Finally, subdividing the essay-writing task, as the authors themselves acknowledge, could improve the interpretability of EEG activity associated with different cognitive processes during writing.

**Second,** given substantial individual differences in writing ability, within-subjects designs may offer more reliable estimates of how LLMs influence cognitive processes. A more systematic comparison of essay quality—using both teacher and AI scoring—would help clarify whether LLM use improves, impairs, or simply alters writing performance.

**Third,** interpretations regarding actual usage patterns of LLMs could benefit from a more natural setting, allowing participants sufficient time to complete the essay. In the current study, participants reported time constraints, which may have led to overreliance on external tools and a potentially distorted view of tool usage.

**Finally,** the results presented in the commented paper appear to reflect attention-oriented differences rather than meaningful variability in learning ability within either behavioral or, especially, EEG outcomes. Thus, task-driven attentional control may substantially influence the reported findings, suggesting reduced mental engagement. However, there is no strong evidence that the LLM use negatively affects performance itself. For example, in the "ability to quote" task, the LLM-group performed significantly worse than the other two groups. Yet, the comparison between the Search-Engine and Brain-Only groups was non-significant ($p = 1$), indicating a null effect. These results contrast with the authors' interpretation that higher task delegation increases "cognitive debt," as the Search Engine group—despite relying on an external tool—showed no impairment.

The authors state that writing an essay without assistance resulted in stronger neural connectivity across all measured frequency bands, with particularly large increases in the theta and high-alpha bands, suggesting that the LLM group engaged fewer of their own cognitive resources during the writing task. However, lower connectivity does not necessarily indicate reduced cognitive engagement. Variability in connectivity (in theta and alpha bands) may indicate differences in attention shifts, changes in strategy, or simply lower motivation. However, cognitive engagement is explained by "degree of investment users make while interacting with digital systems, characterized by factors such as focused attention, emotional involvement and task persistence" (p. 16). Although focused attention was considered, emotional involvement and task persistence received limited discussion, even though these factors may be increased in LLMs (e.g., through reduced frustration). Additionally, while the authors initially introduced the concept of cognitive load and its varying qualities, this framework is not fully integrated into the interpretation of the data. Although the authors provided a compelling and intriguingly consistent narrative regarding tool usage, **many of their conclusions remain speculative, lacking further control experiments to confirm these hypotheses**.

**To conclude**, while we acknowledge the valuable contributions of Kosmyna et al. (2025) in advancing research on the cognitive effects of LLM-assisted writing, **the study would benefit from addressing key methodological concerns and correcting several inconsistencies.** These inconsistencies throughout the manuscript weaken the credibility of the reported findings, including the display of figures that are not described elsewhere in the text and interpretations made without conducting statistical tests. Future research should aim for a more rigorous design, including larger and more diverse samples, clearer operationalizations of cognitive processes, and more detailed statistical reporting. Ultimately, the field stands to gain significantly from further exploration of the cognitive impact of AI tools, but careful attention to these methodological challenges is essential to ensure more reliable and meaningful conclusions.